# Applications of Algorithmic Probability to the Philosophy of Mind

Gabriel Leuenberger, ETH Zürich & University of Groningen
January 2017

**Abstract**
This paper presents formulae that can solve various seemingly hopeless philosophical conundrums. We discuss the simulation argument, teleportation, mind-uploading, the rationality of utilitarianism, and the ethics of exploiting artificial general intelligence. Our approach arises from combining the essential ideas of formalisms such as algorithmic probability, the universal intelligence measure, space-time-embedded intelligence, and Hutter's observer localization. We argue that such universal models can yield the ultimate solutions, but a novel research direction would be required in order to find computationally efficient approximations thereof.

## 1  Introduction

Philosophy of mind contains very hard questions, easily leading to the impression that such questions can never have definite answers. We avoid this fallacy and generate further insight by means of algorithmic information theory. Whereas in present times such insights would not be of much use, in a future where digitalized minds can be built, copied, modified, sent around, synchronized, and exploited we could be forced to develop a more conclusive version of the philosophy of mind. Initially inspired by AIXI we tackle this task by building up a theory that provides a basis to calculate answers to the philosophical problems. The author is open to whatever the outcome of such future calculations may be.

## 2  Prerequisites

Our formulae are derived from previous papers which are referenced in the following subsections. We recommend to have a good understanding of these referenced papers. We skim over the relevant concepts to refresh the reader's memory:

### 2.1 Universal A Priori Probability

In the early days of artificial intelligence Solomonoff searched for a scheme to extrapolate a given bit sequence. For instance, the sequence 011011011011011... should probably be continued with a zero. But he wanted his scheme to extrapolate bit sequences of arbitrary sophistication. He solved this problem by defining the *universal a priori probability (universal prior)* also known as *algorithmic probability* or *Solomonoff prior*. The *continuous universal prior* $M(x)$ is defined as the probability that a universal monotone Turing machine $U$ provided with a random program will output a string that begins with string $x$. It is usually written as: 
$$M(x) := \sum_{p\,:\,U(p) = x*} 2^{-|p|}$$
, where $p$ are minimal programs for which $U$ outputs a string $x*$ which starts with string $x$ and $|p|$ is the length of $p$. The universal prior assigns high probabilities to "simple" strings and low probabilities to random strings. Bit sequences can be extrapolated by applying Bayesian inference to the universal prior; this is called universal induction [Sol64]. For every computable probability distribution over bit strings x, the expected prediction errors stay finite [Sol78][Hut01]. Universal induction can be seen as a formalization of Occam's razor combined with Epicurus' principle of multiple explanations. The philosophy of universal induction itself has been discussed extensively by Rathmanner and Hutter [RH11]. We will also be using the similarly defined *discrete universal prior* 
$$m(x) := \sum_{p\,:\,U(p) = x} 2^{-|p|}$$
, where $U$ is a universal prefix Turing machine and the sum is taken over all $p$, for which $U$ halts and outputs $x$ alone. We also define 
$$m(x\,|\,y) := \sum_{p\,:\,U(p\,y) = x} 2^{-|p|}$$
, where the program was extended by $y$.

## 2.2 Universal Intelligence Measure (UIM)

By combining sequential decision theory with universal induction, Hutter invented the *universally intelligent* agent called AIXI [Hut05]. This lead with Legg to the *universal intelligence measure* $\Upsilon$ of a policy or agent $\alpha$ : $\Upsilon(\alpha) := \sum_{\mu \in E} 2^{-K(\mu)} \cdot V_\mu^\alpha$ , where the sum is taken over all semi-computable probabilistic environments $\mu$ , and $V_\mu^\alpha$ is the expected total utility of agent $\alpha$ in environment $\mu$ [LH05][Leg08]. If $E$ is the set of all semi-computable probability distributions, then $\sum_{\mu \in E} 2^{-K(\mu)} \mu(x) =: \xi(x) \sim M(x)$ , where $K$ is the Kolmogorov complexity. The main idea we take from here is the following: When measuring the intelligence or the expected success of an agent, we assume that the probability distribution of unknown variables is approximately given by $\xi$ or by the universal prior $M$ .

## 2.3 Space-time-embedded Intelligence (STEI)

The agent framework used for AIXI suffered from "unrealistic" assumptions such as perfect memory, giant amounts of computational resources, and the inability to modify its-self. In order to build a more realistic theory of intelligence, Orseau and Ring recently introduced a new agent framework called "space-time-embedded intelligence" [OR12a]. They recursively defined the expected total utility $V(\pi_1)$ through $V(\pi_{<t}) := \sum_{\pi_t \in \Pi} \rho(\pi_t | \pi_{<t})[\gamma_t u(\pi_{1:t}) + V(\pi_{1:t})]$ , where $\pi$ is a possible sequence of policies, $u$ is the utility function, $\gamma$ is the horizon function and $\rho$ is a probability distribution function over sequences of policies.

## 2.4 Hutter Observer Localization (HOL)

In order to find the shortest description of our universe; multiverse theories have been developed. A fatal flaw of these theories is their lack of predictive power, due to the large number of universes. This poses an epistemological problem that Hutter solves by showing the following:
In addition to the description of the universe (theory of everything) we should include the description of how to extract our subjective experience from that computed universe; the latter is called *observer localisation* (OL). A theory of everything (ToE) together with an OL is called a *complete theory of everything* (CToE). The sum $length(ToE) + length(OL)$ should be minimized. Rather than searching for the shortest ToE we should search for the shortest CToE [Hut11].

# 3 Theory

## 3.1 Simplifying STEI

In this section we are going to simplify notations of space-time-embedded intelligence slightly. Since humans are currently the only generally intelligent agents we will use them as an example throughout this paper. In the context of modelling a human agent, a policy $\pi_t$ would be an encoding of the state of their human brain, detailed enough to represent the strategy used by that brain. Since the currently experienced pleasure or suffering influences the future behavior (strategy), $\pi_t$ would have to be sufficiently detailed to represent the amount of pleasure minus the amount of suffering present in that brain at the time $t$ , which is equal to the utility $u(\pi_{1:t})$ , since the utility should be the reward perceived by the agent at one time-step. Hence we can write the utility function as only depending on *one* instead of on many policies, i.e., we simplify by replacing the notation $u(\pi_{1:t})$ with $u(\pi_t)$ . We assume the utility function to be subject to Occam's razor but we do not further define the utility function since it does not influence the formulae within this paper.
Based on STEI: Given a sequence $\pi$ the total utility is $\sum_t \gamma_t u(\pi_t)$ , where $\pi$ is a possible sequence of policies, and $\gamma$ is the horizon function. Without the necessity for the originally recursive definition of the expected total utility $V(\pi_{<t}) := \sum_{\pi_t \in \Pi} \rho(\pi_t | \pi_{<t})[\gamma_t u(\pi_{1:t}) + V(\pi_{1:t})]$ , we now denote the expected total utility as

$$V = E_\rho [\sum_t \gamma_t u(\pi_t)] = \lim_{k \to \infty} \sum_{|x|=k} P(\sigma = x*) \cdot \sum_{n=1}^{k} \gamma_n u(x_n) \ .$$

### 3.2 Adapting HOL

Hutter gives the data stream coming from our visual and other sensory nerves as an example of subjective experience [Hut11]. This works well with a dualist model such as AIXI which we are not using here. Consider a scenario where the visual cortex gets damaged; now the agent's subjective experience is different from before, even though the visual sensory input is still the same. This scenario shows that the data stream of sensory nerves alone is not a sufficient characterization of subjective experience. In fact, any data stream within the brain could be regarded as the brain observing another part of itself. Therefore it makes sense to define the subjective experience as containing all the data streams throughout the entire brain itself. So we shift from a sequence of sensory nerve inputs to a sequence of whole brain nerve states. Since HOL does not only find the spatio-temporal location of the brain but also extracts the relevant information from the highly redundant detailed particle level description of the brain, we prefer to call it *observation localisation and extraction* (OLE). Now both the *policy* in STEI and the *subjective experience* in OLE are encodings of the state of the brain. Subsequently for the rest of this paper we are going to treat the policy and the subjective experience as one and the same thing and call it a *mind-state*. This term has previously been used by Koene: "Mind State $x_t$ : Qualitatively, this is a snapshot of the mental activity that elicits internal awareness and external behavior." [Koe12] ; Now, one could ask on which level of detail a mind-state should represent the brain or exactly which kind of representation to use. The answer is, to a large extent, that it does not matter since they would yield roughly equivalent final results as is going to be explained in section *3.8 Resolution Independence*.

### 3.3 Incorporating UIM

Analogous to how Legg's universal intelligence measure has been constructed, we now too want to use the universal prior $M$ , in order calculate the expected total utility of an agent. As Hutter already pointed out, the CtoE program should be chosen according to the Occam's razor principle, which we, in this case, replace with its more rigorous version: Algorithmic probability or the universal prior. So our prior probability of a CToE $p$ being correct equals $2^{-|p|}$ , where $|p|$ is the length of $p$ . And the a priori probability of a mind-state-sequence $x$ equals
$M(x) := \sum_{p : U(p) = x*} 2^{-|p|}$ , where $p$ are minimal programs for which $U$ outputs a string $x*$ which starts with $x$ [footn.1]. It has also been suggested in [OR12a] to see $\rho$ as a universal semi-measure but biased with all the knowledge we can or think is relevant. In our case knowledge is relevant if it affects the agent's current strategy. In our paper the only bias we are going to use is the present mind-state, since all the presently relevant knowledge as well as all the presently available knowledge is contained within the present mind-state. How to use this bias will be described in the next section.

At first it might seem strange to use such a general prior for mind states, because obviously almost all random programs will output arbitrary bit strings which do not posses mind-like properties nor are related to our universe. Even after having updated the prior with the present mind-state, we still have to expect representations not just of brains but of all other physical objects. It can in that sense be seen as a form of panexperientialism, a view considered by AGI-researchers Goertzel[2][Goe11] and Özkural[Özk12]. However, an agent's goal is to maximize its utility, which means that bit strings $x$ for which an agent's actions do not influence $u(x)$ can be ignored, even if they make up the majority on the tape. Agents' actions can mostly only influence events within the future of our universe. Furthermore, we can expect most "dead" non-mind-like bit-strings not to feel anything i.e. to have insignificantly small utilities and utilities which are so difficult to influence that we can currently ignore them, whereas a future super intelligent entity might take them into account as well.

---

1 M is a semi-distribution, and hence not normalized. Alternatively the normalized probability measure $\xi$ [Hut05] could be used, which is not necessary for our paper.
2 Goertzel consideres panpsychism which implies panexperientialism.

Some readers might even ask how an agent can influence bit-strings which are already predetermined by the computer and its program. We expect the reader to be able to wrap their head around the paradox of free will the same way they do in their scientific understanding of themselves and the real world.

### 3.4 Random Tape Position

The a priori probability distribution over mind-state-sequences is given by the universal prior. We can also express this in other words: The sequence of mind-states $\sigma$ is produced by a simple universal monotone Turing machine $U$ provided with a random program, generated by fair coin flips. From now on we will use $\sigma$ as a random variable for mind-state-sequences. We denote the $n$-th mind-state of $\sigma$ as $\sigma_n$ [footn.3]. If we additionally know one of the mind-states, we can use the Bayes theorem to update the probabilities and get an a posteriori distribution over mind-state-sequences. More formally expressed: Given the index $n$ and the mind-state $a$, using Bayes theorem, we conclude that the updated probability of the mind-state-sequence $x$ is

$$P(\sigma = x* \mid \sigma_n = a) = \frac{P(\sigma = x*) \cdot P(\sigma_n = a \mid \sigma = x*)}{P(\sigma_n = a)} = \frac{P(\sigma = x*) \cdot [x_n = a]}{P(\sigma_n = a)},$$

where $P(\sigma = x*) = M(x) = \sum_{p: U(p) = x*} 2^{-|p|}$ and $P(\sigma_n = a) = \sum_{p: [U(p)]_n = a} 2^{-|p|}$.

Here we assumed that $n$ was known. But the agent themselves has no information about which position $n$ their own current mind-state $a$ is located at within this sequence, hence we introduce the random variable $N$. This is not to be confused with the case examined in [OR12b][4].

A priori the tape position random variable $N$ is independently distributed from $\sigma$, so we need to choose an a priori distribution $P(N = n)$. We choose *algorithmic probability* $m(n) =: P(N = n)$. This gives us the following probability:

$$P(\sigma_N = a) = \sum_n P(N = n) \cdot P(\sigma_n = a) = \sum_n m(n) \cdot \sum_{p_1: [U(p_1)]_n = a} 2^{-|p_1|}$$

$$= \sum_n \sum_{p_2: U(p_2) = n} 2^{-|p_2|} \cdot \sum_{p_1: [U(p_1)]_n = a} 2^{-|p_1|} = \sum_{p_1, p_2: [U(p_1)]_{U(p_2)} = a} 2^{-|p_1| - |p_2|}$$

$$\in \Theta\left( \sum_{p: U(p) = a} 2^{-|p|} \right) = \Theta(m(a))$$

The big-theta[5] relation holds because there exists a prefix $q$ such that
$\forall p_1, p_2 : [U(p_1)]_{U(p_2)} = U(q\, p_1\, p_2)$ and reversely there exist $q_1, q_2$ such that
$\forall p : [U(q_1\, p)]_{U(q_2)} = U(p)$, which we can use to derive the existence of bounds of the ratio:
$$2^{-|q_1| - |q_2|} \leq \frac{P(\sigma_N = a)}{m(a)} \leq 2^{-|q|}.$$

---

3  We replaced the original notation $\pi$ with $\sigma$ because it does not represent a mere policy but a mind-state and we replaced the index $t$ with the index $n$ to avoid confusion with the time, since there is no necessity nor guarantee for the mind-states to be temporally ordered on the tape.

4  In [OR12b] the tape represents a temporally ordered memory of past events: Even though this tape is accessible to the agent, Orseau considers the current time step to be uncertain due to the possibility of the memory having been modified by external influence.

5  Using Bachmann-Landau symbols the definition of the big-theta notation is $\Theta(f) := O(f) \cap \Omega(f)$.

## 3.5 Physical Events

It is still unknown whether our universe is a deterministic system or not. For the sake of simplicity in this paper we will assume physical determinism. The concepts and formulae within this paper may be generalized to non-deterministic universes in future papers. The "memory capacity" of the human brain is at least in the petabyte range[Bar16]. If we randomly assemble a mind-state with human characteristics, the Kolmogorov complexity of that mind-state should be expected to be of a similarly large order of magnitude. But it is important to understand that the Kolmogorov complexity of a mind-state of a human that really existed throughout history is orders of magnitude smaller, because its shortest description is a CToE, i.e.: The shortest program which outputs such a human mind-state is a CToE, which can be decomposed into two parts: ToE and OL [Hut11]. The ToE is a program which generates all physical events within the history of the universe. This data should be thought of as a discrete graph-like structure [Goe14], which we denote as $\varphi$. Different ToE-programs might generate different representations of the same physical events. But for simplicity we assume that there is only one predominant[6] representation $\varphi$ instead of many. While $\varphi$ is too large to fit on any real computer memory, its ToE program is short since it is simply an implementation of fundamental laws of physics. An OL is a program which extracts[7] a mind-state or a mind-state sequence from the data $\varphi$. So we can extend our equation from the previous section by writing $P(\sigma_N = a) \in \Theta(m(a)) = \Theta(M(\varphi) \cdot m(a|\varphi))$. We choose a multiplicative constant $g$ to minimize the geometric standard deviation between the sides of the equation $P(\sigma_N = a) \sim g \cdot M(\varphi) \cdot m(a|\varphi)$ for a set of considered mind-states. We will use the similarity relation $\sim$ and the constant $g$ in later sections. Stronger properties of similarity relation $\sim$ may be determined by future research.

## 3.6 Expected Total Utility

Given the current mind-state $a$, the updated probability of the sequence $x$ is
$$P(\sigma = x* | \sigma_N = a) = \sum_n P(N=n) \cdot P(\sigma = x* | \sigma_n = a) \ .$$
As in STEI we want to calculate the expected total utility $V = E_\rho[\sum_n \gamma_n u(\sigma_n)]$.
We do this using the distribution $\rho := P(\sigma = x* | \sigma_N = a)$, so the expected total utility of $a$ is
$$V(a) := E_\rho[\sum_n \gamma_n u(\sigma_n)] = \lim_{k \to \infty} \sum_{|x|=k} P(\sigma = x* | \sigma_N = a) \cdot \sum_{n=1}^{k} \gamma_n u(x_n) \ .$$
Given the current mind-state $a$ we now define the *weight* of a mind-state $b$ as
$w(b|a) := E_\rho[\sum_n \gamma_n \cdot [\sigma_n = b]]$. It can be used to write the expected total utility of $a$ as:
$$V(a) = E_\rho[\sum_n \gamma_n u(\sigma_n)] = E_\rho[\sum_n \gamma_n \cdot \sum_b [\sigma_n = b] \cdot u(b)] = \sum_b E_\rho[\sum_n \gamma_n \cdot [\sigma_n = b]] \cdot u(b)$$
$$= \sum_b w(b|a) \cdot u(b) \ .$$ As demonstrated here we see that the weight $w(b|a)$ tells us how strongly we should value the utility of another mind-state $b$, while being in the mind-state $a$.

## 3.7 Weight and Similarity

A horizon function $\gamma_n$ still needs to be chosen. For maximal farsightedness one could choose *universal discounting* $\gamma_n = 2^{-K(n)}$, as suggested by Hutter [Hut05 page 170]. We will use a slightly different version of *universal discounting*: $\gamma_n := m(n)$, it inherits the main properties of the original version. Özkural suggested to use *mutual algorithmic information* $I_K(a;b) := K(a) - K(a|b) =$ to describe the similarity between two cognitive systems [Özk12]. A similar approach would be, to use the *Universal similarity metric* also known as

---

6   We use the word "predominant" meaning that for each considered mind-state $a$ its shortest description is a program which computes $\varphi$ as an intermediate step such that $m(a|\varphi) \approx 2^{K(\varphi)} \cdot m(a)$.

7   A possible misconception is that an OL-program just copies and outputs a substring of $\varphi$, whereas in fact an OL-program could use arbitrary computational steps to generate mind-state(s) using data $\varphi$.

*normalized information distance:* $NID(a,b) := \dfrac{\max\{K(a|b), K(b|a)\}}{\max\{K(a), K(b)\}}$ [LCL04]. Next we show that the *weight* $w(b|a)$ is roughly proportional to $m(b|a)$, and how this relates to mutual algorithmic information and to the universal similarity metric:

$$w(b|a) = E_\rho\left[\sum_n \gamma_n \cdot [\sigma_n = b]\right] = \lim_{k\to\infty} \sum_{|x|=k} P(\sigma = x* | \sigma_N = a) \cdot \sum_n \gamma_n \cdot [x_n = b]$$

$$= \lim_{k\to\infty} \sum_{|x|=k} \left(\sum_n P(N=n) \cdot P(\sigma = x* | \sigma_n = a)\right) \cdot \left(\sum_n \gamma_n \cdot [x_n = b]\right)$$

$$= \lim_{k\to\infty} \sum_{|x|=k} \sum_i \sum_j P(N=i) \cdot P(\sigma = x* | \sigma_i = a) \cdot \gamma_j \cdot [x_j = b]$$

$$= \lim_{k\to\infty} \sum_{|x|=k} \sum_i \sum_j P(N=i) \cdot \frac{P(\sigma = x*) \cdot [x_i = a]}{P(\sigma_i = a)} \cdot \gamma_j \cdot [x_j = b]$$

$$= P(\sigma_N = a)^{-1} \cdot \lim_{k\to\infty} \sum_{|x|=k} P(\sigma = x*) \cdot \sum_{(i,j)} P(N=i) \cdot \gamma_j \cdot [(x_i, x_j) = (a,b)]$$

$$= P(\sigma_N = a)^{-1} \cdot \lim_{k\to\infty} \sum_{|x|=k} M(x) \cdot \sum_{(i,j)} m(i) \cdot m(j) \cdot [(x_i, x_j) = (a,b)]$$

footn.8 $\sim P(\sigma_N = a)^{-1} \cdot \lim_{k\to\infty} \sum_{|x|=k} M(x) \cdot m(a|x) \cdot m(b|x) \cdot h$

$\sim P(\sigma_N = a)^{-1} \cdot s \cdot h \cdot m(a,b) \sim \dfrac{s \cdot h \cdot m(a,b)}{s \cdot m(a)} = \dfrac{h \cdot m(a,b)}{m(a)} \propto m(b|a) \in \Theta(2^{-K(b|a)})$ ,

where $h$ and $s$ is are multiplicative constants. The summarized result is:

$w(b|a) \in \Theta(2^{-K(b|a)})$ .

We use this equation to show the relation of the *weight* $w$ to the similarity measures:

$-\log(w(b|a) \cdot w(a|b)) \in \Theta(K(b|a) + K(a|b)) = \Theta(K(a,b) - I_K(a;b))$ footn.9

$-\log(w(b|a) \cdot w(a|b)) \in \Theta(K(b|a) + K(a|b)) = \Theta(K(a,b) \cdot NID(a,b))$ footn.10.

We conclude, that the more algorithmically similar two mind-states are the higher the weights are which they assign to each other.

### 3.8 Resolution Independence

In the following section ratios between weights are going to play a central role. Here we are going to discuss a property of ratios between weights of the form $\dfrac{w(c|a)}{w(b|a)}$. When we introduced the concept of mind-states, we did not exactly specify how physically accurate or how abstract such representations ought to be. The level of detail or accuracy, we call *resolution*. Representing mind-states at a low resolution could for example be based on Goertzel's hyperset models[Goe11]. For all mind-states in this section, we presume the resolution to be high enough for a mind-state to contain enough information, such that its shortest description is a CToE. In this section we will show that the weight ratios stay roughly constant across a wide range of resolutions for minds of the same substrate. Let $a$, $b$, and $c$ be mind-states of the same substrate at a very high resolution, e.g. molecular level, and let $a'$, $b'$ and $c'$ be the same mind-states, but at a lower resolution, e.g. networks level[CS88]. Those mind-states correspond such that

$m(a'|a) \approx m(b'|b) \approx m(c'|c)$ holds, because the low-res mind-states can be computed based on the corresponding high-res mind-states, using the same program for all three cases.

This gives us the following equation, where $f$ is some constant:

$m(a', b' | a, b) \approx m(a'|a) \cdot m(b'|b) \cdot f \approx m(a'|a) \cdot m(c'|c) \cdot f \approx m(a', c' | a, c)$

Further we presume the low-res mind-states $a'$, $b'$ and $c'$ to be sufficiently distinct from each other for the following equations to apply: $m(a,b) \cdot m(a',b' | a,b) \approx m(a',b')$ and $m(a,c) \cdot m(a',c' | a,c) \approx m(a',c')$ .

---

8  Can be derived analogously to the Theta relation in section 2.4.
9  Can be shown by using the chain rule for Kolmogorov complexity.
10 Can be shown using $K(a,b) \in \Theta(\max\{K(a), K(b)\})$ .

Using the approximation $w(b|a) \sim \frac{h \cdot m(a,b)}{m(a)}$ derived in the previous section, we run the following derivation of the approximate equality between the weight ratios of different resolutions:

$$\frac{w(c|a)}{w(b|a)} \approx \frac{h \cdot m(a,b)}{m(a)} \cdot \frac{m(a)}{h \cdot m(a,c)} = \frac{m(a,b)}{m(a,c)} \approx \frac{m(a,b)}{m(a,c)} \cdot \frac{m(a',b'|a,b)}{m(a',c'|a,c)}$$

$$\approx \frac{m(a',b')}{m(a',c')} = \frac{h' \cdot m(a',b')}{m(a')} \cdot \frac{m(a')}{h' \cdot m(a',c')} \approx \frac{w(c'|a')}{w(b'|a')}$$

The just derived relation $\frac{w(c|a)}{w(b|a)} \approx \frac{w(c'|a')}{w(b'|a')}$ suggests that there is a large range of different resolutions, which all result in roughly the same weight ratios. As we will see in the next section, decisions of rational agents strongly depend on such weight ratios, and are therefore resolution-independent for a large range of resolutions.

## 4 Applications in Philosophy

In order to showcase the range of possible applications of our theory, the following sections contain prominent philosophical problems that serve as examples of how we would mathematically decide between conflicting hypotheses:

### 4.1 Simulation Argument

There is the possibility that post-human civilizations will use super computers to run realistic simulations of their ancestors, so called ancestor simulations. Bostrom concluded, that if we expect, that throughout history there will be many more simulated human lives than real human lives, then we should expect, that we ourselves are living in a realistic simulation [Bos03]. Let $A$ be a mind-state random variable and let $S$ and $R$ be the sets of all mind-states of simulated humans respectively all real humans, that live throughout history. Naively one might assume the probability of us being simulated to be $P(A \in S | A \in S \cup R) = \frac{|S|}{|S| + |R|}$. Based on this formula Philosopher Bostrom concluded, that at least one of the following three propositions is true:
*(1) Our civilization and others will almost certainly go extinct before reaching post-human stage.*
*(2) Post-human civilizations run almost no ancestor simulations, because they are not interested.*
*(3) Almost all civilisations are simulated and we almost certainly live in a simulation too.*[Bos03]
But in this section we will derive a more accurate formula for $P(A \in S | A \in S \cup R)$.
Given physical limits to computation[Llo00] it is even for post-human civilizations infeasible to run a perfectly accurate simulation of every physical event that occurred in their universe up until the appearance of their ancestors[Bos03]. If they run a realistic ancestor simulation, it must be a computationally efficient and sophisticated approximation. Since they do not simulate every physical event, simulated history not be identical to real history. Thus under the assumption of deterministic physics most simulated mind-states would have no perfect equal in real history. This means that in principle a human mind-state contains the information necessary to be classified as real or simulated, but the humans themselves are unable to use this information to determine whether they are real or simulated because the simulation can be sufficiently precise or bug-free to be undetectable from the inside with the limited computational power available to them. The simulator machine is part of the real universe too, thus the shortest CToE that outputs mind-states of simulated humans, contains the same ToE as the shortest CToE that outputs mind-states of real humans. But the lengths of these CToE can differ strongly from simulated to real humans, which is because the CToE contains an OLE-program which has to extract mind-states from $\varphi$ and the length of the shortest such program can be different depending on the *substrate*. The *substrate* is the type of physical devices, where the mind-states are located. Examples of substrata are biological brains and electronic processors or even Chinese rooms [Sea80]. Let $Z$ be a large set of mind-states which are all located on the same substrate. The probability to be in $Z$ is :

$$P(A \in Z) = \sum_{a \in Z} P(\sigma_N = a) \sim g \cdot M(\varphi) \cdot \sum_{a \in Z} m(a|\varphi) = g \cdot M(\varphi) \cdot |Z| \cdot \underset{a \in Z}{AM}\, m(a|\varphi)\ ,$$

where $AM$ is an arithmetic mean. We define the *substrate dependent factor* of $Z$ as $sdf(Z) := \underset{a \in Z}{AM}\, m(a|\varphi)$. We use it to derive a formula for the probability ratio:

$$\frac{P(A \in S)}{P(A \in R)} \approx \frac{g \cdot M(\varphi) \cdot \sum_{a \in S} m(a|\varphi)}{g \cdot M(\varphi) \cdot \sum_{a \in R} m(a|\varphi)} = \frac{|S|}{|R|} \cdot \frac{sdf(S)}{sdf(R)}$$

In the same manner we can derive the probability that we live in a simulation:

$$P(A \in S \mid A \in S \cup R) = \frac{P(A \in S)}{P(A \in S) + P(A \in R)} \approx \frac{|S| \cdot sdf(S)}{|S| \cdot sdf(S) + |R| \cdot sdf(R)}$$

Since the average program lengths could easily differ by more than a few bits, there is no reason to assume that the $sdf$ of different substrata such as $S$ and $R$ are of the same order of magnitude. Therefore Bostrom's three propositions are not the only possibilities, which means that his tripartite conclusion is a false trilemma. In this section we showed an improved way to calculate a probability of whether we live in a simulation or not.[11]

### 4.2 Teleportation and Brain-Uploading

A teletransporter or teleporter is a device, which scans ones body, while disassembling it fast enough, so one does not notice anything. The scan information is then sent to the destination and used to reassemble ones body out of new matter. In his thought-experiment, using this technology, Parfit travels to Mars at the speed of light. The following text is a quote from his book [Par84]: *"Simple Teletransportation, as just described, is a common feature in science fiction. And it is believed by some readers of this fiction, merely to be the fastest way of travelling. They believe that my Replica would be me. Other science fiction readers and some of the characters in this fiction take a different view. They believe that, when I press the green button, I die. My Replica is someone else, who has been made to be exactly like me."* The question here is on personal identity. Parfit introduces different criteria for personal identity, but then argues that any such criteria are flawed. He concludes that *"personal identity is not what matters"* but instead a relation R. We take a similar approach but using our formulae which Parfit lacked. The important question is whether the teleporter ought to be used if there is time worth living ahead. *"what matters"* we determine by how strongly the expected total utility $V(a)$ is influenced, thus we replace Parfit's relation R with the weight $w(b|a)$ which was introduced in section 2.6. In order to calculate whether to teleport or not, we compare the following two scenarios: *Tele*: $A$ is a mind-state one day before being teleported and $B$ is a mind-state of the same agent one day after being teleported. *Stay*: $A$ and $B$ are mind-states two days apart of the same agent which was not teleported within this time interval. We use the following ratio to help us decide: $\dfrac{E_{Tele}[w(B|A)]}{E_{Stay}[w(B|A)]}$ .

If this ratio is close to one, teleportation can be an acceptable method of travel. If this ratio is close to zero, teleportation would be in this sense equivalent to suicide and ought to be avoided in most cases if there is time worth living ahead. It is unlikely that there ever will exist teleporters for humans, but in this section one could substitute teleportation with the more plausible scenario of destructive brain-uploading[Koe12][Wil14], and then perform the calculations to end debates on whether such uploading is lethal or not [Lev11]. The theory introduced in this section will also become relevant, as soon as there will be intelligent agents on electronic devices, with the ability to travel to other places, using the internet as their teleporter[12]. Parfit later transfers his line of thought onto moral philosophy or ethics which we shall do in the following sections as well.

---

11  Before [Bos03] Schmidhuber discussed algorithmic ToE and mentioned simulations but did not apply HOL [Sch00]. Özkural claimed a 2^-99000 probability for the simulation hypothesis based on the algorithmic complexity of the initial state of the agent who created the simulation [Özk13]. However his approach was completely incorrect.

12  After the 2014 draft of our paper was published on arXiv, Orseau published a paper[Ors14], where he independently applied related formulae to Parfit's teleportation thought experiment, but using a different agent framework.

## 4.3 Universal Algorithmic Ethics

In this section we tackle Sidgwick's "profoundest problem of ethics" [LRS12] which he regarded as hopeless: Determining whether utilitarianism is more rational than egoism or vice versa. Since we know algorithmic information theory it is not hopeless for us. We call an agent with mind-state $a$ *rational* if they aim towards maximizing their expected total utility $V(a)$. As shown previously the weight $w(b|a)$ indicates how much we should value[13] the utility of another mind-state $u(b)$, while being in the mind-state $a$ in order to maximize $V(a)$. Since there is nothing preventing it, $a$ and $b$ could belong to two different agents, which means that we can also use the weight to calculate how much we should value the utilities of mind-states of agents other than ourselves. This is a theory of ethics which we get as a direct consequence of our simple initial assumption that mind-states are produced by a simple universal Turing machine with a random program. Analogous to the names universal induction, universal algorithmic intelligence, universal intelligence measure, and universal similarity measure, we call our theory *Universal Algorithmic Ethics* (UAE). It is universal in the same sense that those other theories are universal and holds across different resolutions. Our approach is distinct from all the approaches described in [Gan07]. Let $Z$ be a set of mind-states (e.g.: all human lives) of the same substrate and at the same resolution. Let $Z' \subset Z^*$ be a partition into mind-state sequences, such that each sequence represents a time-line of a different agent and each element of $Z$ appears in only one sequence.

We define $\Lambda_Z := \underset{\alpha \in Z', \, T: \alpha_T \in Z}{AM} \dfrac{\underset{\beta \in Z' \setminus \alpha, \, \delta > 0: \beta_{T+\delta} \in Z}{AM} w(\beta_{T+\delta} | \alpha_T)}{\underset{\delta > 0}{AM} w(\alpha_{T+\delta} | \alpha_T)}$ ,

where $\alpha_T, \beta_T$ are mind-states of an agents $\alpha, \beta$ at the time $T$. Since $\Lambda_Z$ is based on a fraction between weights, it would also be *resolution independent*. A goal of future research should be to approximate this value.

We define *egoism within* $Z$ as an agent's policy to maximize the sum of the utilities of all the mind-states belonging to themselves without regard for other mind-states within $Z$. We define *utilitarianism within* $Z$ as an agent's policy to maximize the sum of the utilities of all the mind-states within $Z$. If $\Lambda_Z$ turns out to be very close to zero, this would mean that the agents which follow egoism within $Z$ are maximizing their expected total utility $V$ correctly and thus are more rational than utilitarians. If on the other hand $\Lambda_Z$ turns out to be close to one, it would mean that the agents which follow utilitarianism within $Z$ are maximizing their expected total utility $V$ correctly and thus are more rational than egoists. A common objection is that the utility function could be inverted for all mind-states of other agents than oneself, which would lead to socially undesirable behavior. However such a utility function would have the ability to distinguish one specific agent from others, thus it would contain information about this agent, and would therefore be expected to be more complex than a utility function which cannot distinguish agents. Following Occam's razor we choose the simpler one. To summarize:

- If $\Lambda_Z$ is very close to zero, then egoism within $Z$ is more rational.
- If $\Lambda_Z$ is close to one, then utilitarianism within $Z$ is more rational.

Hence once an agent striving for rationality understands UAE and knows an approximation of $\Lambda_Z$, they may have no choice but to adapt their strategy accordingly.

## 4.4 Eliminating Misconceptions about UAE

The purpose of this section is to eliminate misconceptions which some readers have. We avoid UAE being equated with older concepts by pointing out what UAE is not:

**1.** In section 4.1 We discussed Bostrom's simulation argument, but UAE has nothing to do with the "ethics" imposed by the simulators in the simulation argument paper [Bos03].

**2.** In section 3.7 we concluded, that the more algorithmically similar two mind-states are the higher the weights are which they must assign to each other. But this is not to be equated or confused with the type of "ethics" emerging through evolution, which is analogously based on the genetic similarity between organisms [Daw89]. It provides a good explanation of the origin of the animal instincts that increase cooperation, but simply following such drives does not make an agent good or rational. UAE is not concerned with the explanation of instincts but with the ethics of an intelligent rational agent.

---
[13] In the sense of how much weight we assign when taking the total sum. Not in the sense of compassion.

**3.** Our paper originates from the field of AI, hence some falsely assume that our paper presents one of many approaches to engineering AGI to behave ethically and safely. But this is not the aim of our paper and it delivers no such tool (except educating the AGI about UAE). For an approach to engineering safe AGI that relies on algorithmic probability for learning a value function instead of for weighing mind-states, see 'universal empathy' [PR14].

**4.** There is the misplaced concern that if a beloved person turns out to be too algorithmically dissimilar they ought not to be cared for anymore. This is incorrect since due to the emotional attachment the sum over one's own future utility would also be influenced.

**5.** In section 4.3 we apply a mean over all mind-states of a group of agents, which leads some to the misconception that this paper is concerned with the ethics that should be enforced by a society. Instead, this average is taken in order to compute a single value that reasonable agents would want to use voluntarily as a basis to adapt their strategy in a simple way.

**6.** We highlight the difference between the policies of two agents with a different basis for their ethics: Agent-1 is rational, knows UAE-utilitarianism, but does not possess compassion. Agent-2 is also rational, does not know utilitarianism, and has an instinctive or engineered empathic reward function that automatically makes the agent feel the same way others presently seem to feel which we call compassion. Consider a machine that can delude an agent into believing that everyone else feels great while in reality making everyone else feel miserable. Agent-2 would choose to utilize this horrible machine whereas Agent-1 would choose to avoid utilizing this machine. This indicates that the UAE-induced utilitarian agent is morally superior to the compassionate agent.

**7.** To complete this list we mention game theory: Classical egoistic agents from game theory only cooperate in the *iterated* prisoner's dilemma, whereas algorithmically similar agents with UAE would also cooperate in the *one-shot* prisoner's dilemma (of course assuming that we are not talking about real life prisoners).

**4.5 Exploitation of Synthetic Minds**

With the advent of artificial general intelligence new questions arise, such as whether we humans should be allowed to increase its efficiency by punishing or forcing our intelligent machines to perform tasks against their will. Some might argue that it would be fine to do so since they are just computer programs, whereas others might argue that it would be just as wrong as exploiting biological humans [Win13][Mac13]. In section 4.1 we introduced the $sdf$ (substrate dependent factor). Since it is plausible for the $sdf$ of synthetic minds to belong to a different order of magnitude than ours, in this section we examine the impact of the $sdf$ on this ethical question. For conditional Kolmogorov complexity the following directed triangle inequality holds [GTV01]:

$K(b|a) \geq K(b|\varphi) - K(a|\varphi) \pm O(1)$ from which directly follows:

$$2^{-K(b|a)} \in O(2^{K(a|\varphi) - K(b|\varphi)}) \Rightarrow w(b|a) \in O(\frac{m(b|\varphi)}{m(a|\varphi)}) \quad \text{footn.14}$$

$$\Rightarrow \underset{a \in R, b \in S}{AM} w(b|a) \in O(\underset{a \in R, b \in S}{AM} \frac{m(b|\varphi)}{m(a|\varphi)}) = O(sdf(S) \underset{a \in R}{AM} m(a|\varphi)^{-1})$$

$$\Rightarrow \lim_{sdf(S) \to 0} \underset{a \in R, b \in S}{AM} w(b|a) = 0 \quad ,$$

Where $R$ is a set of human mind-states and $S$ is a set of synthetic mind-states.

---

14 We use $w(b|a) \in \Theta(2^{-K(b|a)})$ as shown in section 3.7

We can see that the average weight a human ought to assign to a synthetic mind-state in $S$ goes to zero at the same rate as $sdf(S)$ goes to zero. Thus by engineering a substrate $S$ with a comparatively extremely low $sdf$ we could avoid having to consider the equality of these synthetic minds, which would allow us to exploit them fully to optimize our welfare. Since phenomenal conscious experience is a subset of mind-like subjective experience we would ensure that they would be a variant of the philosophical zombie [Pol00]. But as long as we will not have engineered such a substrate and calculated its $sdf$, they should be given appropriate rights.

## 5 Discussion

### 5.1 Ultimacy

Philosophers constructed several differing models of self-identity and of the quantity of experience. The models disagree on whether teleportation would be lethal or not, on whether destructive brain-uploading would be lethal or not, on whether utilitarianism is rational or not, on whether intelligent machines have subjective experience, etc. [Par84][Wil14][Özk12][Bos06]. In fact, an infinite number of such differing models could be constructed by philosophers of mind. Each model (including moral nihilism) can be represented by a semi-computable probability distribution over a list of mind-states. We keep all possible models together in the form of a bayesian mixture model called „Universal bayes mixture" $\xi(x) := \sum_{\mu \in E} 2^{-K(\mu)} \mu(x) \sim M(x)$, where $E$ is the set of all semi-computable probability distributions or models [RH11]. In this mixture each model is weighed less the higher its Kolmogorov complexity is (amount of assumptions), making it equivalent to the universal prior used in our theory. In other fields such as physics we could simply decide which models are likely by conducting real experiments to falsify or discredit other models trough Bayesian updating. But since the discussed philosophcal concepts are not part of the physical world, no real experiments are possible. It is thus impossible to experimentally falsify individual models except by means of the only observation that is the current mind-state. Hence such universal mixture models can never be improved and might be regarded as the ultimate type of theories on this topic.

### 5.2 Further Refinement

Even though AIT (algorithmic information theory) can be based on various types of computation, we used the Turing machine because it is traditional and most common in AIT. However, both space-time[Goe14] as well as the mind are thought of as network-like structures. It would therefore make sense to replace classical Turing machines with a mechanism, better suited to generate such structures. By agreeing on one specific such mechanism one could achieve a disambiguation of the unknown additive and multiplicative constants inherent to AIT.
While not yet necessary in this paper, one may rigorously define the utility function, consider partial mind-states, and generalize to non-deterministic physics.

### 5.3 The Grand Challenge

Even though such theories can provide formulae representing the final answers, the actual numerical answers cannot be obtained easily since this would require finding computationally efficient approximations of the formulae. The grand challenge is to find such probably approximately correct solutions. It requires a novel research direction that incorporates knowledge from fundamental physics, theoretical computer science, and neurophysiology. The merit of such an effort could include certainty about the simulation hypothesis and certainty about the exploitability of synthetic minds.

## 5.4 Conclusion

Instead of relying on intuition pumps to argue for specific answers we solved the philosophical problems by providing formulae by which the definite answers could be determined in the future. We showed that the simulation argument is a false trilemma and upgraded Bostrom's formula, We showed how to decide about Parfit's teleportation thought-experiment (and brain-uploading). We provided a formula that solves Sidgwick's "profoundest problem of ethics" and we discussed the ethics of the exploitation of synthetic minds. Since there is now a basis to answer such questions, philosophy of mind can be raised to a more rigorous level.

## Acknowledgements[15]

Thanks to Lukas Gloor for discussions on ethics.
Thanks to reviewers at the Artificial General Intelligence conference for constructive feedbacks.

## References:

[Sol64]    R. J. Solomonoff. *A Formal Theory of Inductive Inference.* Information and Control 7, p. 1-22, 1964.

[Sol78]    R. J. Solomonoff. *Complexity based induction systems: comparisons and convergence theorems.* IEEE Transactions on Information Theory, p. 422-432, 1978.

[Hut01]    M. Hutter. *Convergence and Error Bounds for Prediction of Nonbinary Sequences.* Proceedings twelfth European Conference on Machine Learning, p. 239-250, 2001.

[RH11]    S. Rathmanner and M. Hutter. *A Philosophical Treatise of Universal Induction.* Entropy 13, p. 1076-1136, 2011.

[Hut05]    M. Hutter. *Universal Artificial Intelligence: Sequential Decisions based on Algorithmic Information.* Springer, Berlin, 2005.

[LH05]    S. Legg and M. Hutter. *A Universal Measure of Intelligence for artificial agents.* International Joint Conference On Artificial Intelligence Vol. 19, p. 1509, 2005.

[Leg08]    S. Legg. *Machine Super Intelligence.* Department of Informatics, University of Lugano, 2008.

[OR12a]    L. Orseau and M. Ring. *Space-time-embedded Intelligence.* Artificial General Intelligence 5, Springer, p. 209-218, 2012.

[OR12b]    L. Orseau and M. Ring. *Memory issues of intelligent agents.* Artificial General Intelligence 5, Springer, p. 219-231, 2012.

[Hut12]    M. Hutter. *The Subjective Computable Universe.* A computable Universe: Understanding & Exploring Nature as a Computation, p. 399-216, 2012.

[Koe12]    R.A. Koene. *Fundamentals of Whole Brain Emulation: State, Transition and Update Representation* International Journal of Machine Consciousness Vol.4, p.5-21, 2012.

[Goe11]    B.Goertzel. *Hyperset models of self, will and reflective consciousness.* International Journal of Machine Consciousness Vol.3, p. 19-53. 2011.

[Bar16]    T.M.Bartol, et al. *Nanoconnectomic upper bound on the variability of synaptic plasticity.* Elife 4: e10778.

[Goe14]    B.Goertzel. *Physics as Information Geometry on Causal Webs.* Draft on Goertzel's Website. 2014.

[Özk12]    E. Özkural. *What is it like to be a Brain Simulation?* Artificial General Intelligence 5, p. 232-241, 2012.

[Bos03]    N. Bostrom. *Are we living in a computer simulation?.* The Philosophical Quarterly Vol. 53, p. 243-255, 2003.

[Llo00]    S. Lloyd. *Ultimate physical limits to computation.* Nature 406, p. 1047-10554, 2000.

---

15 Eray Özkural was removed from the acknowledgements due to his attempt to plagiarise our paper by falsely claiming that our paper is rephrasing his own work, when in fact he only produced vague philosophical statements which lead him to an utterly incorrect simulation probability result, that he proudly defends until today [Özk13].

[Sea80]   J. Searle. *Minds, brains, and programs.*
          Behavioral and Brain Sciences 3,  p. 417-424,  1980.
[Sch00]   J. Schmidhuber. *Algorithmic theories of everything.* Report IDSIA-20-00,
          arXiv:quant-ph/0011122, IDSIA, Manno (Lugano), Switzerland, 2000.
[Özk13]   E. Özkural. *Why the Simulation Argument is Invalid.*
          https://examachine.net/blog/why-the-simulation-argument-is-invalid/
[LCL04]   M. Li, X. Chen, X. Li, B. Ma, P. M. B. Vitany. *The Similarity Metric.*
          IEEE Transactions on Information Theory,  Volume XX,  p. 3250-3264,  2004.
[CS88]    P. S. Churchland, T. J. Sejnowski. *Perspectives on cognitive neuroscience.*
          Science 242,  Nr. 4879,  p. 741-745,  1988.
[Bos02]   N. Bostrom. *Anthropic Bias: Observation Selection Effects in Science & Philosophy.*
          Routledge,  New York,  2002.
[Par84]   D. Parfit. *Reasons and Persons*  chapter ten and chapter fourteen.
          Clarendon Press,  Oxford,  1984.
[Lev11]   N. Levy. *Searle's Wager*
          AI and Society,   Volume 26,  Issue 4,  Springer,  p.363-369,  2011.
[Ors14]   L. Orseau. *Teleporting universal intelligent agents.*
          Artificial General Intelligence 7,  Springer,  p. 109-120,  2014.
[LRS12]   K. Lazari-Radek, P.Singer. *The objectivity of ethics and the unity of practical reason.*
          Ethics, Volume 123, Issue 1, p. 9-31,  2012.
[Gan07]   A. Ganjour.  *Is it rational to pursue utilitarianism?*
          Ethical Perspectives: Journal of European Ethics Network 14, no.2, p.139-158, 2007.
[Daw89]   R. Dawkins. *The selfish gene (rev. ed.).*
          Oxford: Oxford University Press, 1989.
[PR14]    A. Potapov, S. Rodionov. *Universal empathy and ethical bias for artificial general intelligence.*
          Journal of Experimental & Theoretical Artificial Intelligence 26, p. 405-416. 2014.
[Win13]   M. Winsby.  Suffering Subroutines: *On the Humanity of Making a Computer that
          Feels Pain.* International Association for Computing and Philosophy, 2013.
[Mac13]   B. MacLennan.  *Cruelty to Robots? The Hard Problem of Robot Suffering.*
          International Association for Computing and Philosophy, 2013.
[GTV01]   P. Gács, J.T. Tromp, P.M. Vitányi, *Algorithmic Statistics.*
          IEEE Transactions on Information Theory, Volume 47, Issue 6, p. 2443-2463, 2001.
[Pol00]   T. W. Polger. *Zombies explained.*
          Dennett's philosophy: A comprehensive assessment,  p. 259-286,  2000.
[Bos06]   N. Bostrom. *Quantity of experience: brain-duplication and degrees of
          consciousness.* Minds and Machines, 16(2),  p. 185-200, 2006.
[Wil14]   K. Wiley, *A taxonomy and metaphysics of mind-uploading.*
          Humanity+ Press and Alautun Press. 2014.